\documentclass[10pt,twocolumn]{article} 

\usepackage{cite}
\usepackage{amsmath,amssymb,amsfonts}
\usepackage{algorithm}
\usepackage{algpseudocode}
\usepackage{graphicx}
\usepackage{textcomp}
\usepackage{multirow}    
\usepackage{booktabs}    
\usepackage{xcolor}
\usepackage{wrapfig}
\usepackage[margin=0.75in]{geometry}
\usepackage{authblk}

\def\BibTeX{{\rm B\kern-.05em{\sc i\kern-.025em b}\kern-.08em
    T\kern-.1667em\lower.7ex\hbox{E}\kern-.125emX}}
\usepackage{hyperref}

\newcommand{\keywords}[1]{\par\vspace{6pt}\noindent\textbf{Keywords:} #1}

\newcommand\length{\texttt{l}}
\newcommand\width{\texttt{w}}

\newcommand\asp{\texttt{asp}}

\newcommand{\MSE}{\texttt{MSE}}
\newcommand{\HRRP}{HRRP }

\begin{document}
\title{MFN Decomposition and Related Metrics for High-Resolution Range Profiles Generative Models\\
\thanks{Accepted at RadarConf25. This work was performed using HPC resources from GENCI-IDRIS (Grant 2024-AD011014422R1).}
}

\author[1,2]{Edwyn Brient}
\author[1]{Santiago Velasco-Forero}
\author[2]{Rami Kassab}
\affil[1]{Center for Mathematical Morphology (CMM), Mines Paris, PSL University, Fontainebleau, France\\\texttt{santiago.velasco@minesparis.psl.eu}}
\affil[2]{Advanced Radar Concepts (ARC), Thales Land and Air Systems, Limours, France\\\texttt{rami.kassab@thalesgroup.com}}
\date{}

\maketitle

\begin{abstract}
High-resolution range profile (\HRRP) data are in vogue in radar automatic target recognition (RATR). With the interest in classifying models using HRRP, filling gaps in datasets using generative models has recently received promising contributions. Evaluating generated data is a challenging topic, even for explicit data like face images. However, the evaluation methods used in the state-of-the-art of HRRP generation rely on classification models. Such models, called ``black-box", do not allow either explainability on generated data or multi-level evaluation. This work focuses on decomposing HRRP data into three components: the mask, the features, and the noise. Using this decomposition, we propose two metrics based on the physical interpretation of those data. We take profit from an expensive dataset to evaluate our metrics on a challenging task and demonstrate the discriminative ability of those.  
\end{abstract}


\section{Introduction}
\emph{High-Resolution Range Profile} (\HRRP) represents the projection of the target's echoes along the line of sight (LOS) of the radar, compressing radar data into small one-dimensional representations. This dimensionality reduction allows real-time processing on radar systems, including machine learning models. However, this reduction also reduces each data's information level while saving the target's structural features. 

The main aim of integrating embedded models in radars is target recognition. In this context, recent works studied the involvement and performance of different machine learning algorithms on HRRP data classification \cite{bauw2020unsupervised, sun}. Various authors have studied the architecture of classification models using a single HRRP as input \cite{hrrp_clf1,hrrp_clf2,hrrp_clf3}. However, we can hardly imagine extending those studies to a large-scale database. The variability of \HRRP data (Fig. \ref{fig:sensitivity}) and the diversity of \HRRP features of a ship depending on its aspect and depression angles turn this task into an ``ill-posed" problem. 

Some works tackled the \HRRP generation task to improve classifiers by filling gaps in the dataset's scenarios \cite{multi_aspect_gen, one_shot_gen, recog_aware}. In these studies, the performance of models is calculated through classifiers' accuracy. They demonstrate the generation ability of their models by comparing the accuracy of a classifier trained on both true data and generated ones with the accuracy of the classifier trained only on the true data. 

This habit values the application of generative models in a radar context, yet it does not allow the comparison between generative models because authors do not share the same classifier. Also, many biases can help a classifier achieve satisfactory accuracy without extracting an object's features from data. For example, \cite{maritime_radar} shows that radar signal suffers from diverse effects during rainy weather. These effects transform \HRRP data, resulting in a bias in data captured during wet weather compared to a calm one. Finally, using a black-box algorithm to demonstrate the performance of a model hinders the explainability of generative models. \label{bias}

We propose the first \HRRP decomposition to unlock explainability both for classification and generation, named Mask Features Noise (MFN) decomposition, based on the three components defining this decomposition. Especially in a generation context, this decomposition allows for the evaluation of specific features of the generated data. For example, we used the decomposition in \cite{MOI} to evaluate the ability of our generative model to generate \HRRP data with the correct number of cells activated by the target depending on the aspect angle. This work also provides new metrics, especially designed to be discriminative depending on the target. Such a metric adapted to our task could be used in metric learning, a promising framework for the challenging task of classifying targets based on their \HRRP \cite{zhong_contrastive_2023}. The code of both our decomposition and the related metrics is available at \url{https://github.com/EdwynBrient/MFN-Metrics/}.

Our contributions can be summarized as follows:
\begin{itemize}
    \item We provide the first physical based decomposition designed for \HRRP data, allowing the training of a classification model not influenced by noise bias.
    \item We propose two metrics especially built for their discriminative ability in the context of target identification.
    \item Both the decomposition and the metrics can be directly used in a learning context, we provide their Pytorch code.
\end{itemize}

\section{data and previous work}
\subsection{Data}

A radar detects the echo of its emitted signal, which—after processing steps \cite{richards2005fundamentals}—can be represented as a two-dimensional radial grid of amplitudes $\sigma$, known as the radar cross section (RCS). A high amplitude indicates that an object has reflected the signal back toward the radar. This echo is characterized by two axes: $r$, the range (distance from the radar), and $\theta$, the azimuth angle. A range profile is obtained by integrating $\sigma$ tangentially over a detection cone spanning the angular interval from $\Theta$ to $\Theta + \Delta\Theta$. The difference between two consecutive range values defines the radar's range resolution, denoted as $\Delta r$. The grid is limited to s range cells. Although the radar also has an angular resolution in $\theta$, this resolution is embedded in the amplitude of the High Resolution Range Profile after transformation and is no longer explicitly visible.
\label{eq:basics}
\begin{equation}
    \text{hrrp}(r_i) = \sum\limits_{\theta \in \Theta + \Delta\Theta} \sigma(r_i, \theta_i)
\end{equation}

Some geometrical phenomena originate from the projection of echoes along the LOS of the radar. Two angles mainly describe the acquisition scenario geometry : the \emph{aspect} angle and the \emph{depression angle}. The aspect angle ($asp$) describes the way the target is observed by the radar based on target's heading ($hdg$) and the radar's azimuth ($az$) : $asp = hdg-az$. Additionally, the depression angle is the angle formed between the horizontal and the radar's LOS.

\begin{figure}[h]
    \centering
    \includegraphics[width=1\linewidth]{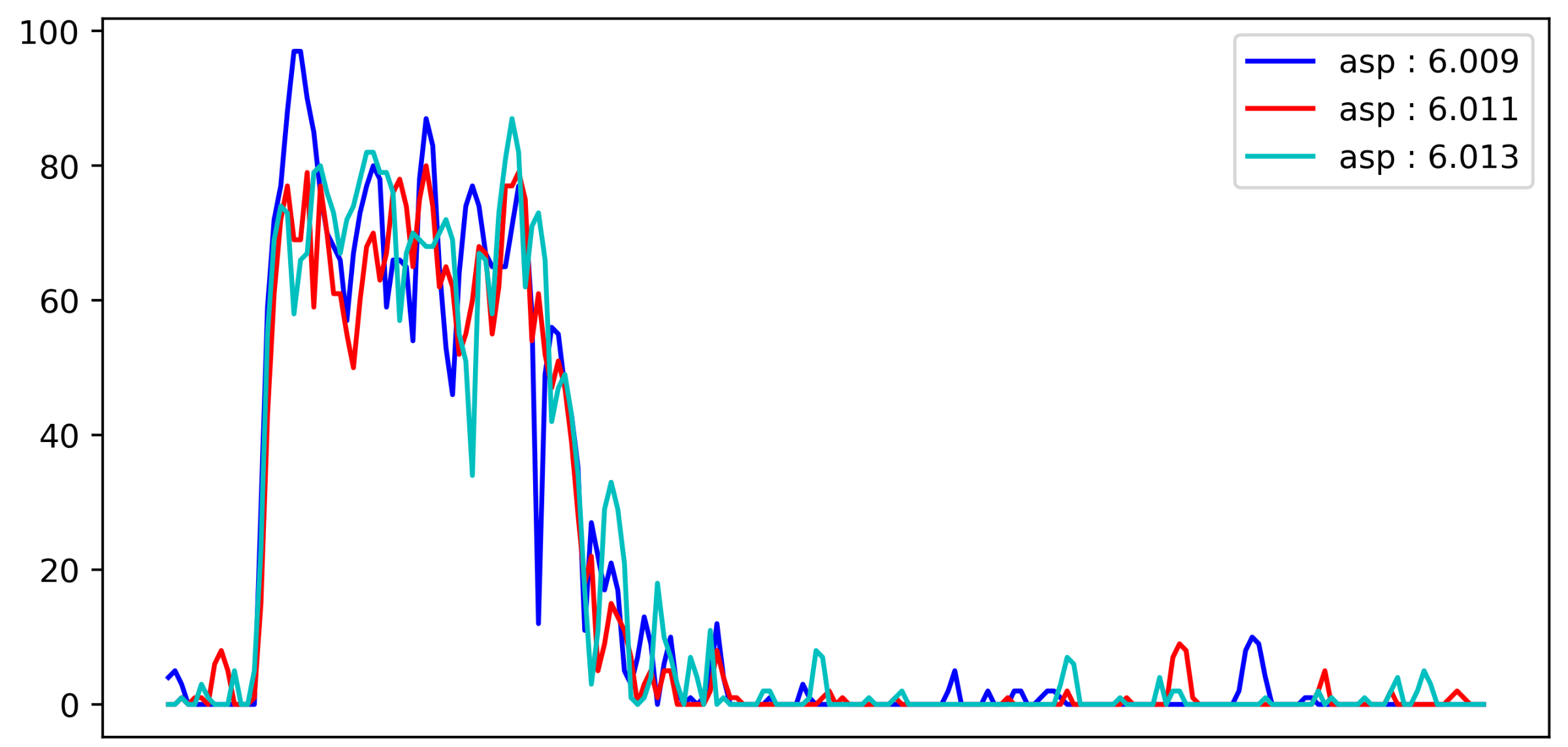}
    \caption{Three HRRPs of the same ship captured the same day at very close aspect angles. \textbf{X-axis:} range cells, \textbf{Y-axis:} amplitude.}
    \label{fig:sensitivity}
\end{figure}

We leverage an extensive radar HRRP database containing 753 vessels and more than 900.000 signals to validate the decomposition and metrics we introduce. We couple \textit{Automatic Identification System} (AIS) signal and the radar database to retrieve meta data such as the aspect angle and Maritime Mobile Service Identity (MMSI) of each data. This variable is a unique serial number defining ships. Our ground radar is nearly sea level and the target's distance from the radar is around kilometers. Thus, we consider the impact of depression angle negligible. However, the aspect angle describes our theoretical scenarios. The nature of HRRP data makes it very sensitive. Even on low clutter data (Fig. \ref{fig:sensitivity}), the same ship can produce different HRRP representations at close aspect angles. 

\subsection{Previous Work}
\label{prev_work}
\begin{figure}[h]
    \centering
    \includegraphics[width=1\linewidth]{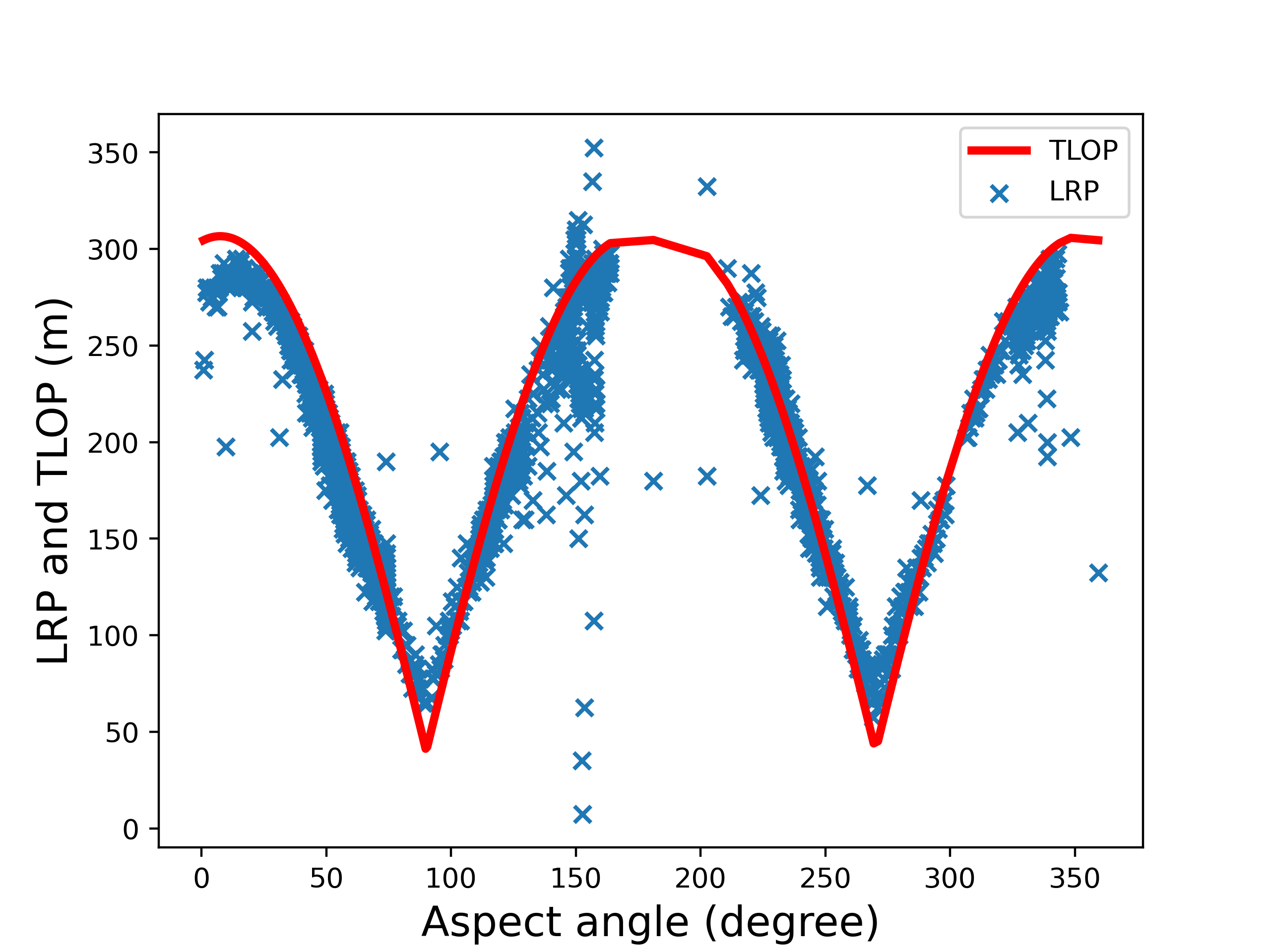}
    \caption{Example of TLOP and LRP of multiple data at various aspect angles. \textbf{Red} : TLOP, theoretical projection length of a rectangle on the LOS of the radar. \textbf{Blue} : LRP calculated on the data depending on the aspect angle. Each cross corresponds to a data. }
    \label{fig:tlop_lrp}
\end{figure}

Prior work \cite{MOI} has explored generation of HRRP data conditionnally to the dimensions of the ship and the aspect angle. To validate the generation ability of our models, we introduced a method to determine the length of an object on HRRP data. First, we define the \textit{cells of interest} (COI): the cells transmitted by the return of the radar signal reflecting on the target. To select those cells, we apply a uniform filter on the signal then use a threshold of 50\% of the maximum amplitude of the data. We finally use a morphological closing operator to fill gaps smaller than 14 cells in object’s representation. We do not normalize our data because we believe the amplitude is a feature of our targets. Otherwise, we could have used the global maximum, so this definition is not limiting. We can now define the empirical length of the objects on HRRP data, called Length on Range Profile (LRP). Using the previous definition of cells of interest, we compute a binary mask of COI in data. In these masks, COI cells are set to 1 while every other cells are 0. The LRP of a HRRP is then the length (in cell) of the first object in a HRRP. From the projection process resulting in HRRP data, we deduce the Theoretical Length of Object Projection (TLOP) on these data: 

\begin{equation}
    TLOP(\asp, \length, \width) = \length |cos(\asp)| + \width |sin(\asp)|
\end{equation}

with \asp \hspace{1pt} the aspect angle and \length (resp. \width) the length (resp. width) of the ship. Note that this formula is an approximation of the shape of a ship by a rectangle. However, it can be adapted depending of the expected target. Fig. \ref{fig:tlop_lrp} shows the LRP and TLOP of a ship captured at various aspect angles covering 360°. We select a large ship captured around 8.000 times for visualization. Besides some outliers, the LRP scatters fit the TLOP. 

\begin{figure*}[h]
    \centering
    \vspace{-10pt}
    \includegraphics[width=0.8\linewidth]{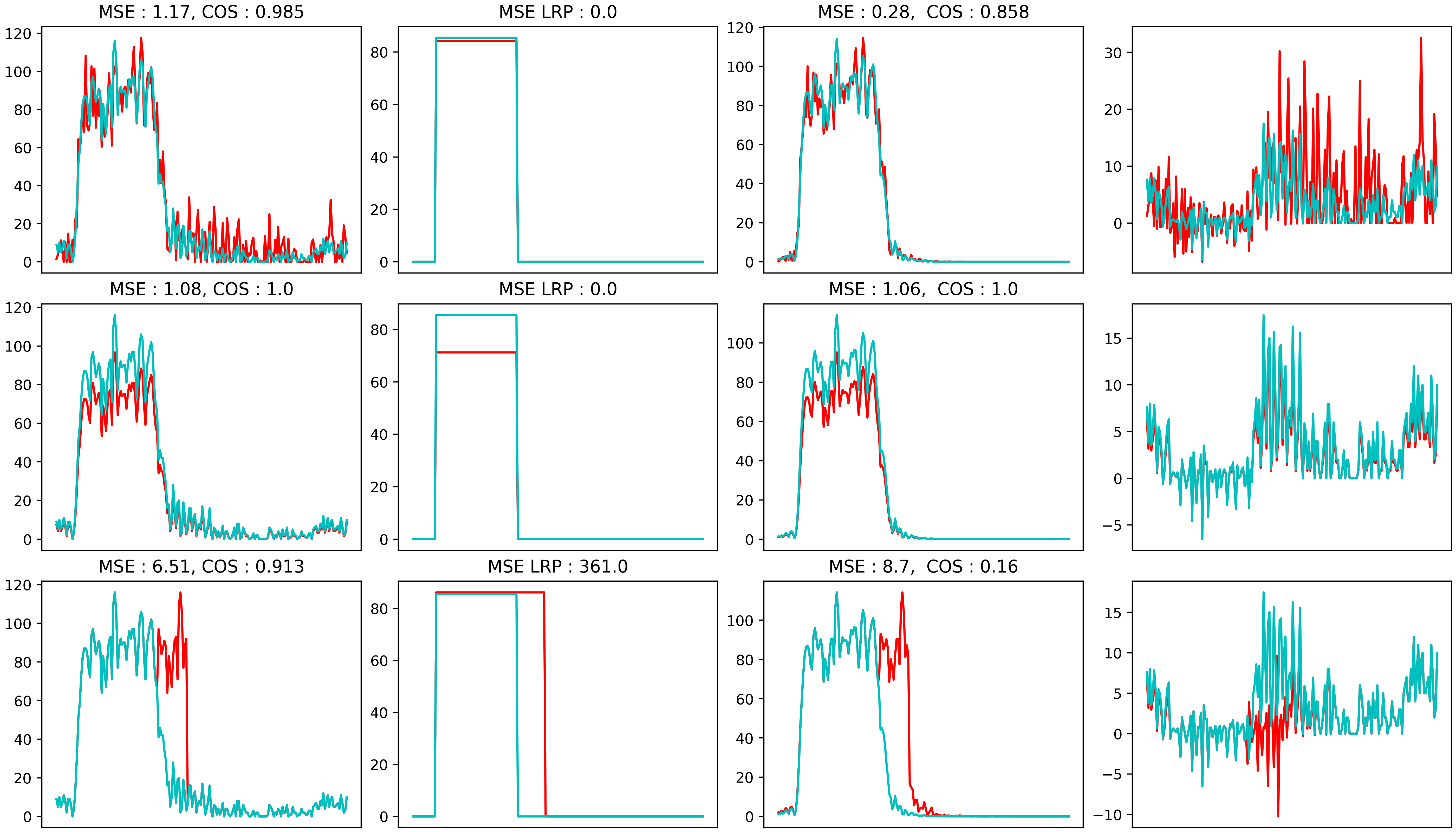}
    \caption{Examples of decomposition (with $\sigma = 0.5$) with comparison metrics between the standard data (\textbf{cyan}) and transformed data (\textbf{red}). From left to right: raw data, \textit{m} components, \textit{f} components, and \textit{n} components. \textbf{Top}: Adding noise to the data transforms slightly the \textit{m} and \textit{f} component, while the \textit{n} component captures most of the noise. \textbf{Middle}: The transformed data is the standard data at lower amplitude. The MSE metric captures this difference, while the cosine similarity is the same for both data. \textbf{Bottom}: We add low-frequency components to the standard data. Our cosine similarity is way more discriminant than the standard one without decomposition.}
    \label{fig:ex_decomp}
\end{figure*}

We introduced new metrics to avoid evaluating our models' generative ability using a black-box model. Aware of the sensitivity of HRRP data, these metrics were designed to be flexible. Let $m$ be a metric that has to be maximized to get the best performance. We want to compare a data $x_g$ with a set of data $(x)_i$, our metrics were designed as follow : 

\begin{equation}
    top_m = \max\limits_{\asp_i \in [\asp_g - 5\text{\textdegree}, \asp_g + 5\text{\textdegree}]} m(x_g, x_i)
\end{equation}

with $\asp_g$ (resp. $\asp_i$) the aspect angle of data $x_g$ (resp. $x_i$). A data is then no longer compared with another one but with a set of data, factoring then the variability of this kind of data. Although this flexibility helps evaluating generated data, the issue of bias in Section \ref{bias} is not treated with this method. Therefore, we propose in this paper a decomposition to liberate these radar data from the noise skewing the metrics.

\section{Decomposition}


Multiscale decomposition involves representing a signal in terms of different frequency bands or scales, allowing the extraction of fine and coarse details. Our decomposition turns HRRP signals into three components at three different scales. The \textit{m} component stands for the mask of the ship in data. Then, the \textit{f} component represents the robust HRRP features that are common for several representations of an object for a given aspect angle. Finally, the \textit{n} component combines clutters and high-frequency features that are not considered as robust. 

We introduced a method to segment the object in HRRP data \ref{prev_work}. This segmentation is a key to discriminating the object features from noise. We thus segment the objects' area in HRRP using the cells of interest. The straightforward operation to select the features from this segmentation is to apply a mask on data at the object location. However, the masked HRRP is not representative of the object's features because of the huge variation between two close data (fig. \ref{fig:sensitivity}). To obtain robust representation of ships in our decomposition, we need to keep structural features for given aspect angle while losing small variations that could vanish from a signal to another. Thus, we add a Gaussian filter (with standard deviation $\sigma$) in our pipeline to smoothen features and keep multi-cells features. This way, the step 6 of algorithm \ref{alg:mfn_decomp} separates high-frequency component (added to the noise component \textit{n}.) from low-frequency component \textit{f}. The parameter $\sigma$ allows to refine the definition of low and high frequency depending on the resolution $\Delta r$ of the radar. The residual is then a combination of clutter and non-robust peaks/through in HRRP. The symbol $\odot$ corresponds to Hadamard product. Finally, smooth\_with\_distance refers to a function that smooths its first argument (a vector v) based on its distance from a cell of v with the closest activated cell of a mask (coi) ($v[i] \propto v[i] \times \exp(-2 \times \mathrm{dist}(i, \text{coi}))$).

\begin{algorithm}
\caption{MFN Decomposition Algorithm}
\label{alg:mfn_decomp}
\begin{algorithmic}[1]
\State \textbf{Input:} High Resolution Range Profile data $RP$, variance of Gaussian filter $\sigma$
\State \textbf{Initialization:} Initialize $m$, $f$, $n$ as tables of length $s$
\Statex
\State $coi \gets$ mask of cells of interest of $RP$
\State $m \gets mean(RP \odot coi) \times coi$
\State $m_f \gets$ smooth\_with\_distance(m, coi)
\State $f \gets gaussian\_filter(RP \odot m_f, \sigma)$ 
\State $n \gets RP-f$
\State \textbf{return} $m$, $f$, $n$
\end{algorithmic}
\end{algorithm}

This algorithm is deep learning compatible. An example of utilization could be in target recognition. The utilization of f and the width of m component as input for a classification model could prevent the noise from impacting the predictions. Thus, the bias in noise should vanish while using both the object's LRP and features during training. Also, to evaluate a generative model, we could decompose generated samples to evaluate separately the features, the noise and the LRP. 

More generic methods such as EMD or wavelets already exist, however the components we isolate here come from physics behind data. We indeed know the projection of our object on LOS is only a part of the HRRP data. The remaining cells are a mixture of different clutters independent from the object or without pattern. This feature in HRRP data is crucial and separates clutter from object's features. 

Our decomposition can be used as a testing metric, for overall performance out of the training scope. Yet, it can also be turned into training losses. All transformations from raw data to components are differentiable. Thus, depending on the generation goal, the loss can focus on diverse combination of components. We propose in next section two metrics designed for HRRP generation and metric learning.

\section{Metrics}

\subsection{Standard Metrics}
We select two metrics to compare data using decomposition. First, a universal yet controversial metric to compare two data is the \emph{Mean Squared Error} (\MSE). This metric gives an overall distance between two data, yet, it suffers from scale difference and is uniformly weighted on its standard formulation. For example if we compared a generated data with a measured data containing two ships, the \MSE \hspace{1pt} would be huge even if the generation has a good fidelity. Second, the cosine similarity is scale invariant and gives a better score of match between data features. Combined, the \MSE \hspace{1pt} gives an overall score Radar Cross Section (RCS) dependant and the cosine similarity gives a RCS invariant similarity score between features. 

We recall these metrics formulas: 
\begin{equation}
    \MSE(x_1, x_2) = \left\| x_1 - x_2 \right\|^{2}
\end{equation}
\begin{equation}
    dist_{\cos}(x_1, x_2) = \frac{x_1 \cdot x_2}{\left\|x_1\right\| \cdot \left\|x_2\right\|}
\end{equation}
with $x_1$ and $x_2$ two data, $\cdot$ the dot product and $\|\cdot\|$ a norm.  

\subsection{Proposed Metrics}
\label{subsec:proposed_metrics}
Each of these metrics suffers from bias and limitations on both the regular HRRP data and the \textit{f} component of the decomposition. No matter whether we use decomposition, the \MSE \hspace{1pt} between two data of a small ship will be lower than that of bigger vessels. The clutter is hard to predict because it is stochastic, but its amplitude is way lower than the amplitude within cells of interest. Thus, an error on the clutter (composing more cells in data of small ships) is lower than an error on cells of interest. In a machine learning context, we do not want a generative model to focus the results on a subclass of ships unless we explicitly ask for it. To remove this bias in \MSE \hspace{1pt} metric, we propose a new version depending on the LRP of data : 
\vspace{-5pt}
\begin{equation}
    \MSE_f(x_1, x_2) = \frac{\MSE(f_1, f_2)}{\frac{LRP_1+LRP_2}{2}}
\end{equation}
with $f_1$ (resp. $f_2$) the \textit{f} component of data $x_1$ (resp. $x_2$) and $LRP_1$ (resp. $LRP_2$) the LRP (introduced in \ref{prev_work}.) of data $x_1$ (resp. $x_2$).

With only positive amplitude in our HRRP cells, the discriminative ability of a cosine similarity applied directly on data is decreased compared to other applications. Moreover, a HRRP data is composed of many clutter cells that are not containing any information to discriminate data. In order to solve both issues, we introduce a cosine similarity applied on the union of cells of interest of both data. To formulate this, let $I$ be the set of s cells composing our HRRPs. Using the \textit{m} component, we can define $COI_1$ and $COI_2$ the subsets of $I$ containing cells of interest of data. Let $f_1$ and $m_1$ (resp. $f_2$ and $m_2$) denote the f and m components of $x_1$ (resp. $x_2$).

\begin{equation}
    \cos_f(x_1, x_2) = \frac{\sum_{i \in \mathcal{I}} (f_{1norm})_i \times (f_{2norm})_i}{\sqrt{\sum_{i \in \mathcal{I}} (f_{1norm})_i^2} \cdot \sqrt{\sum_{i \in \mathcal{I}} (f_{2norm})_i^2}}
\end{equation}

with $f_{1norm}$ (resp. $f_{2norm}$) the normalized \textit{f} component, i.e $f_1-mean(f_1 \odot m_1$) (resp. $f_2 - mean(f_2 \odot m_2)$) of data $x_1$ (resp. $x_2$). For readability, we use $\mathcal{I} = COI_1 \cup COI_2$ to describe the union on targets' cells of interest. We introduce validation experiments and results in the following section.

\section{Experiments and Results}

\subsection{Experiments}
\label{subsec:exp}
To validate our decomposition and metrics, we test our metrics on different ships at different aspect angles. The standard metrics are already discriminative if the ships and aspect angle are too far (Table \ref{tab:no_const}). Consequently, we select pairs of ships with close length, with a tolerance of five meters. The data we use to validate the metrics are thus from ships with similar length within a bin of 10\textdegree. Using this method, we select 10 pairs of vessels for each of the 36 bin with 30 data within the aspect angle bin. Overall, we compare 10.800 different data from multiple ships of more than 50 meters to have enough features for comparison. 

Following the intuition explained in \ref{prev_work}, in all experiments besides the first, we select the top metrics between one data and a set of data. Within our framework, we provide the top scores between each data of a ship and the 30 data of the other within each bin. The discriminative ability is then the difference between the top scores of data from the same ship and data from different ships. To give a visual interpretation of these scores depending on the aspect angle, we plot an average score of our top metrics on each bin. We recall one of the goal of our decomposition is to give a representation isolating the noise. Thus, our proposed metrics are in competition with biased metrics. 

\subsection{Results}

\begin{figure*}[h]
    \centering
    \vspace{-10pt}
    \includegraphics[width=1\linewidth]{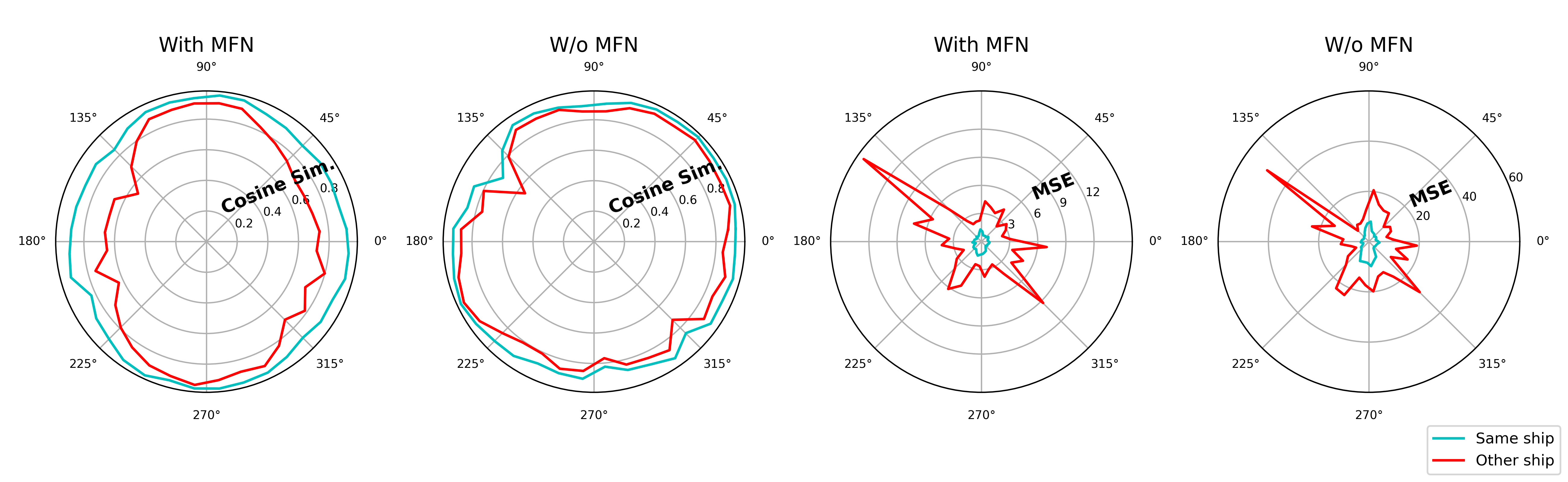}
    \caption{Top metrics comparison. \textbf{Cyan}: top metrics between a data and a set of data from the same ship. \textbf{Red}: top metrics between data from different ships. We note a larger difference with our decomposition.}
    \label{fig:top_wwo_mfn}
\end{figure*}

We first show the results of our metrics on the framework described in last subsection. To demonstrate the importance of comparing metrics on vessels of similar lengths, we begin with an experiment on random ships (without length constraint). The table \ref{tab:no_const} contains top and mean metrics on HRRP data with the decomposition (columns MFN.) and with no transformation (columns Std.). We use the normalized \MSE \hspace{1pt} \ref{subsec:proposed_metrics} on both types of data to ensure similar scale with and without decomposition. 

\begin{table}[h]
    \centering
    \setlength{\tabcolsep}{8pt}
    \renewcommand{\arraystretch}{1.2}
    
    \begin{tabular}{l|cc|cc}
        \hline
        \multirow{2}{*}{Metric} & \multicolumn{2}{c|}{\textbf{Top}} & \multicolumn{2}{c}{\textbf{Mean}} \\
        \cline{2-5}
        & MFN & Std & MFN & Std \\
        \hline
        \MSE same &   0.60    &    3.30   &    2.77   &    9.22   \\
        \MSE diff &   17.07    &   89.73    &   23.46    &   108.0    \\
        Cos same &   0.93    &    0.91   &    0.81   &    0.91   \\
        Cos diff &   0.59    &    0.62   &    0.38   &    0.62   \\
        \hline
    \end{tabular}
    \vspace{0.1cm}
    \caption{Results on ship selected with no length constraint.}
    \vspace{-0.3cm}
    \label{tab:no_const}
\end{table}

The limitation of a cosine similarity applied directly to the range profile stands in this table. In this context, there is no difference between the top and standard cosine similarities. However, all metrics with and without decomposition (especially the \MSE.) are discriminative. We also note a significant difference between the highest and average measurements, showing the presence of data modes within the angle bins. We focus from now on the most challenging pairs to demonstrate the proposed metrics' effectiveness. 

Fig. \ref{fig:top_wwo_mfn} provides the results of the framework we describe in last subsection \ref{subsec:exp}. We compare the metrics computed on data captured on the same vessel with data captured on two targets. The more significant the difference between values of each bin the more discriminative the metric. 

An aspect angle of 0\textdegree or 180\textdegree corresponds to a ship with radial heading. In such scenarios, the LRP is maximized compared to other aspect angle's scenarios (Fig. \ref{fig:tlop_lrp}). Thus, comparing two data at those aspect angles involves more cells of interest than at 90\textdegree or 270\textdegree. As a result, the difference of cosine similarity between data from the same target and between data from two targets is greater horizontally than vertically on the Fig. \ref{fig:top_wwo_mfn} because the component \textit{f} is projected in a greater space (more COI for aspect angles 180\textdegree and 0\textdegree.).

\begin{figure}[h]
    \centering
    \includegraphics[width=0.9\linewidth]{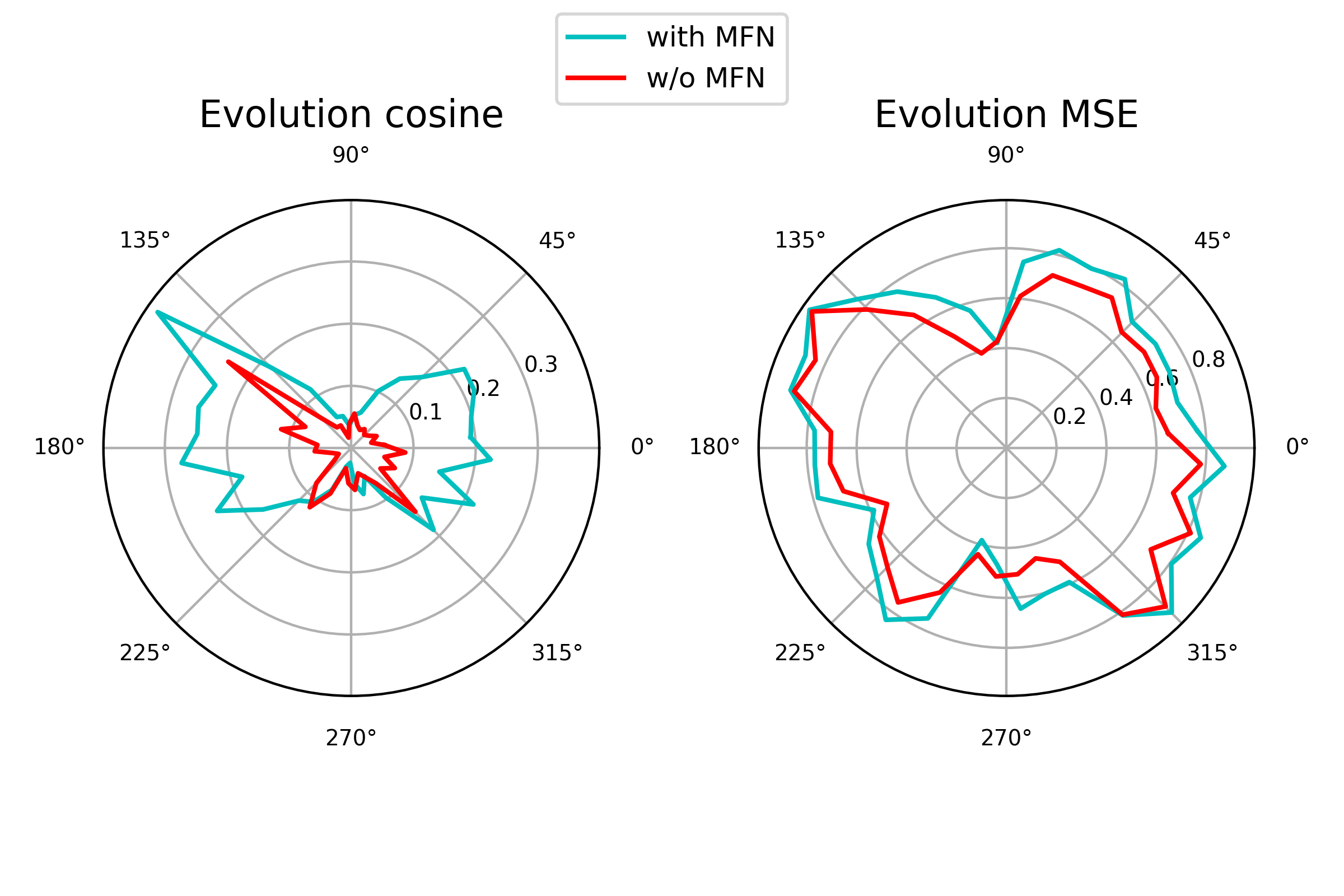}
    \caption{Relative evolution for both metrics. \textbf{Cyan}: metrics introduced in \ref{subsec:proposed_metrics} using the decomposition. \textbf{Red}: standard metrics, we recall that we normalized the standard \MSE \hspace{1pt} by the mean LRP of targets.}
    \label{fig:evo_metrics}
\end{figure}

We value metrics that discriminate pairs of data acquired from different targets compared to those acquired on the same target. Thus, we calculate the relative evolution for all aspect angle's bin to evaluate those. The evolution of our cosine similarity is greater than the classical cosine similarity (Fig. \ref{fig:evo_metrics}) on almost every aspect angle bins. The proposed \MSE \hspace{1pt} is virtually better everywhere, whereas the standard \MSE \hspace{1pt} is biased on raw HRRP data. 

\begin{figure}[h]
    \centering
    \includegraphics[width=0.9\linewidth]{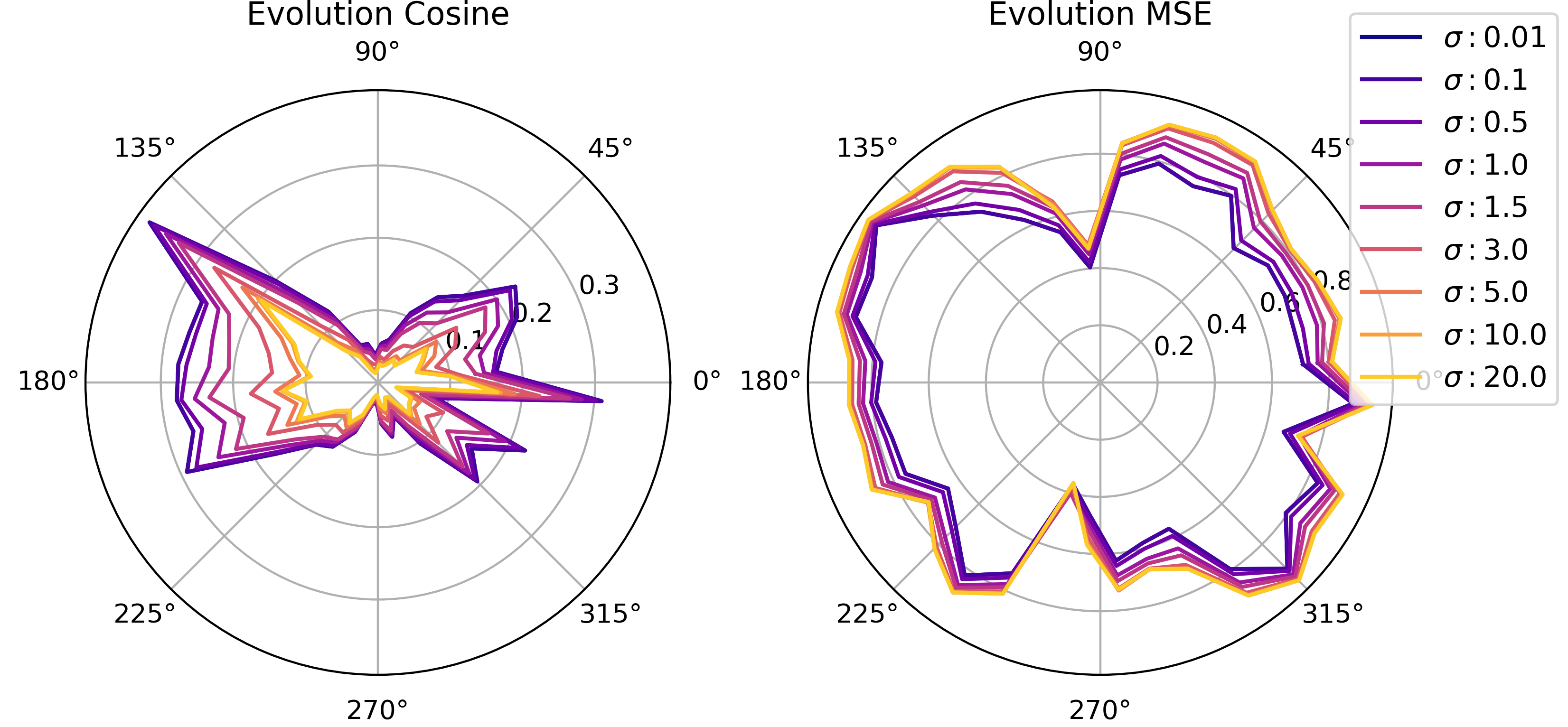}
    \caption{Relative evolution depending on the variable sigma of the Gaussian filter (algo. \ref{alg:mfn_decomp}).}
    \label{fig:evo_sigma}
\end{figure}

Finally, we reproduce the Fig. \ref{fig:evo_metrics} on multiple scales $\sigma$ of the MFN Gaussian filter to compare the evolution of our metrics depending on this variable. The greater the sigma, the smoother the \textit{f} components. On the one hand, the evolution of our cosine similarity gets worse with greater sigma because \textit{f} components lose feature diversity between targets (fig. \ref{fig:evo_sigma}). On the other hand, with sigma growing, the \textit{f} component becomes an approximation of the overall RCS of the ship, which is a good discriminant between ships. However, the \textit{f} component loses all its robust features if $\sigma$ gets too high. We select $\sigma = 0.5$ from this figure for the other experiments because it seems to be a good balance between consistent cosine similarity and \MSE. 
\vspace{-5pt}
\subsection{Discussion}
Our improvements in metrics evolution show promising steps to evaluate similarities between HRRP data. The Gaussian filter is the first proposition for isolating robust features from others. However, a cell of interest with high amplitude will be considered more robust than one with low amplitude using our decomposition. The notion of robustness for HRRP data is complex because of the nature of those. 

The LRP and \textit{m} component depend on the definition of cell of interest. A good definition of a cell of interest depends on the database. We tried to use the 
Constant False Alarm Rate (CFAR) method \cite{skolnik2001radar} to define those cells, but the noise level of our data is too different from one data to another. 

We hope these contributions will be the first step in separating the evaluation of HRRP's generative models from classifiers. Reproducibility in the defense field is a tough topic, yet being able to compare different methods is a mandatory point in research. 

\vspace{-5pt}

\section{Conclusion}
We provide the first physical-based decomposition of HRRP data, allowing explainability in classification and generation tasks. This decomposition bypasses the noise bias in HRRP target recognition by isolating objects' robust features within data. We propose new metrics based on this decomposition that go beyond standard metrics limitations on HRRP data. Using a large-scale database and a challenging validation framework, we demonstrate the discriminative ability of those metrics depending on the aspect angles of targets.

\bibliographystyle{unsrt}
\bibliography{biblio}

\begin{thebibliography}{10}

\bibitem{bauw2020unsupervised}
Martin Bauw et~al.
\newblock From unsupervised to semi-supervised anomaly detection methods for
  {HRRP} targets.
\newblock In {\em RadarConf20}, pages 1--6. IEEE, 2020.

\bibitem{sun}
Lin Sun, Jianpo Liu, Yuanqing Liu, and Baoqing Li.
\newblock Hrrp target recognition based on soft-boundary deep svdd with lstm.
\newblock In {\em 2021 International Conference on Control, Automation and
  Information Sciences (ICCAIS)}, pages 1047--1052, 2021.

\bibitem{hrrp_clf1}
Chih-Lung Lin, Tsung-Pin Chen, Kuo-Chin Fan, Hsu-Yung Cheng, and Chi-Hung
  Chuang.
\newblock Radar high-resolution range profile ship recognition using
  two-channel convolutional neural networks concatenated with bidirectional
  long short-term memory.
\newblock {\em Remote Sensing}, 13(7), 2021.

\bibitem{hrrp_clf2}
Yujia Diao, Shuowei Liu, Xunzhang Gao, Afei Liu, and Zhiwei Zhang.
\newblock {CNN} based on multiscale window self-attention mechanism for radar
  {HRRP} target recognition.
\newblock In {\em 2022 7th ICSIP}, pages 281--285, 2022.

\bibitem{hrrp_clf3}
Bin Xu, Bo~Chen, Jinwei Wan, Hongwei Liu, and Lin Jin.
\newblock Target-aware recurrent attentional network for radar {HRRP} target
  recognition.
\newblock {\em Signal Processing}, 155:268--280, 2019.

\bibitem{multi_aspect_gen}
Yiheng Song, Qiang Zhou, Wei Yang, Yanhua Wang, Cheng Hu, and Xueyao Hu.
\newblock Multi-view hrrp generation with aspect-directed attention gan.
\newblock {\em IEEE Journal of Selected Topics in Applied Earth Observations
  and Remote Sensing}, 15:7643--7656, 2022.

\bibitem{one_shot_gen}
Liangchao Shi, Zhehan Liang, Yi~Wen, Yihong Zhuang, Yue Huang, and Xinghao
  Ding.
\newblock One-shot {HRRP} generation for radar target recognition.
\newblock {\em IEEE Geoscience and Remote Sensing Letters}, 19:1--5, 2022.

\bibitem{recog_aware}
Yue Huang, Yi~Wen, Liangchao Shi, and Xinghao Ding.
\newblock Recognition-aware hrrp generation with generative adversarial
  network.
\newblock {\em IEEE Geoscience and Remote Sensing Letters}, 19:1--5, 2022.

\bibitem{maritime_radar}
Xun Wang, Menghan Wei, Ying Wang, Houjun Sun, and Jianjun Ma.
\newblock Radar signal behavior in maritime environments: Falling rain effects.
\newblock {\em Electronics}, 13(1), 2024.

\bibitem{MOI}
Edwyn Brient, Santiago Velasco-Forero, and Rami Kassab.
\newblock Conditional generation of hrrp data for unseen ships using generative
  models and ais information.
\newblock {\em Submitted}, 2025.

\bibitem{zhong_contrastive_2023}
Yijin Zhong, Wei Lin, Ying Xu, Lexing Huang, Yue Huang, and Xinghao Ding.
\newblock Contrastive {Learning} for {Radar} {HRRP} {Recognition} {With}
  {Missing} {Aspects}.
\newblock {\em IEEE Geoscience and Remote Sensing Letters}, 20:1--5, 2023.

\bibitem{richards2005fundamentals}
M.A. Richards.
\newblock {\em Fundamentals Of Radar Signal Processing}.
\newblock McGraw-Hill Education (India) Pvt Limited, 2005.

\bibitem{skolnik2001radar}
M.~I. Skolnik.
\newblock {\em Introduction to Radar Systems}.
\newblock McGraw-Hill Book Company, USA, 2001.

\end{thebibliography}

\end{document}